\documentclass[11pt,a4paper]{article}
\usepackage[hyperref]{acl2020}
\usepackage{times}
\usepackage{latexsym}

\usepackage{amsmath,graphicx,amsfonts}
\usepackage{float}
\usepackage{color, soul}
\usepackage{multirow}
\usepackage{booktabs}
\usepackage{pifont}
\newcommand{\cmark}{\ding{51}}%
\newcommand{\xmark}{\ding{55}}%
\usepackage[normalem]{ulem}

\usepackage{microtype}

\aclfinalcopy 

\title{Enforcing Encoder-Decoder Modularity in Sequence-to-Sequence Models}
\author{Siddharth Dalmia,$^{1}$\thanks{Work done at Facebook AI Research.} Abdelrahman Mohamed,$^2$ Mike Lewis,$^2$ Florian Metze,$^{1}$ Luke Zettlemoyer$^2$ \\
$^1$Carnegie Mellon University $^2$Facebook AI Research \\
\texttt{sdalmia@cs.cmu.edu, abdo@fb.com}}

\date{}

\begin{document}

\maketitle
\begin{abstract}
Inspired by modular software design principles of independence, interchangeability, and clarity of interface, we introduce a method for enforcing encoder-decoder modularity in seq2seq models without sacrificing the overall model quality or its full differentiability. We discretize the encoder output units into a predefined interpretable vocabulary space using the Connectionist Temporal Classification (CTC) loss. Our modular systems achieve near SOTA performance on the 300h Switchboard benchmark, with WER of $8.3\%$ and $17.6\%$ on the SWB and CH subsets, using seq2seq models with encoder and decoder modules which are independent and interchangeable.
\end{abstract}

\section{Introduction}
\label{sec:intro}
Modularity is a universal requirement for large scale software and system design, where ``a module is a unit whose structural elements are powerfully connected among themselves and relatively weakly connected to elements in other units.''~\cite{Baldwin_99}. In addition to independence, good software architecture emphasises interchangability of modules, a clear understanding of the function of each module, and a unambiguous interface of how each module interacts with the larger system. In this paper, we demonstrate that widely adopted seq2seq models lack modularity, and introduce new ways of training these models with independent and interchangeable encoder and decoder modules that do not sacrifice overall system performance. 

Fully differentiable seq2seq models \cite{chan2016listen,bahdanau2016end, bahdanau2014neural, vaswani2017attention} play a critical role in a wide range of NLP and speech tasks, but fail to satisfy even very basic measures of modularity between the encoder and decoder components. The decoder cross-attention averages over the continuous output representations of the encoder and the parameters of both modules are jointly optimized through back propagation. This cause a tight coupling, and prevents a clear understanding of the function of each part. As we will show empirically, current seq2seq models lack modular interchangability, i.e. retraining a single model with different random seeds will cause the encoder and decoder modules to learn very different functions, so much so that interchanging them radically degrades overall model performance. Such tight coupling makes it difficult to measure the contributions of the individual modules or transfer components across different domains and tasks.

In this paper, we introduce a new method that guarantees encoder-decoder modularity while also ensuring the model is fully differentiable. We constrain the encoder outputs into a predefined discrete vocabulary space using the connectionist temporal classification (CTC) loss \cite{graves2006connectionist} that is jointly optimized with the decoder output token-level cross entropy loss. This novel use of the CTC loss ensures discretizing the encoder output units while respecting their sequential nature.  By grounding the discrete encoder output into a real-world vocabulary space, we are able to measure and analyze the encoder performance.  We present two proposals for extending the decoder cross-attention to ingest probability distributions, either using probability scores of different hypotheses or using their rank within a fixed beam. Combining these techniques enables us to train seq2seq models that pass the three measures of modularity; clarity of interface, independence, and interchangeability. 

The proposed approach combines the best of the end-to-end and the classic sequence transduction approaches by splitting models into grounded encoder modules performing translation or acoustic modeling, depending on the task, followed by language generation decoders, while preserving full-differentiability of the overall system. We present extensive experiments on the standard Switchboard speech recognition task. Our best model, while having modular encoder and decoder components, achieves a competitive WER $8.3\%$ and $17.6\%$ on the standard 300h Switchboard and CallHome benchmarks respectively.

\section{Baseline Seq2seq models}
\label{sec:current}
\subsection{Attention-Based encoder-decoder models}
Conditioned on previously generated output tokens and the full input sequence, encoder-decoder models \cite{sutskever2014sequence} factorize the joint target sequence probability into a product of individual time steps. They are trained by minimizing the token-level cross-entropy~(CE) loss between the true and the decoder predicted distributions. Input sequence information is encoded into the decoder output through an attention mechanism \cite{bahdanau2014neural} which is conditioned on current decoder states, and run over the encoder output representations.  
\subsection{Models optimized with the CTC loss}
\label{sec:ctc}
Rather than producing a soft alignment between the input and target sequences, the Connectionist Temporal Classification (CTC) loss \cite{graves2006connectionist} maximizes the log conditional likelihood by integrating over all possible monotonic alignments between both sequences.
\begin{align*}
    Y &=  \texttt{Softmax} \left( \texttt{Encoder}(X)*W_o \right)\\
    \mathcal{F}_{CTC}(L,Y) &= -\log \sum_{z \in \mathcal{Z}(L,T)} \left(\prod_{t=1}^{T} Y_{z_t}^t \right)
\end{align*}
Where\footnote{We decided to omit the discussion of the extra CTC blank symbol in the above equation for clarity of presentation, \cite{graves2006connectionist} provides the full technical treatment of the subject} $W_o \in \mathbb{R}^{d \times |V_e|}$ projects the encoder representations into the output vocabulary space, $L$ is the output label sequence, $\mathcal{Z}(L,T)$ is the space of all possible monotonic alignments of $L$ into $T$ time steps, and the probability of an alignment $z$ is the product of locally-normalized per time step output probabilities $Y \in \mathbb{R}^{T \times |V_e|}$. The forward-backward algorithm is used for efficient computation of the marginalization sum. Only one inference step is required to generate the full target sequence in a non-autoregressive fashion through the encoder-only model.

\subsection{Joint CTC and Attention-Based models}
In \cite{suyounkim, Karita2019ACS}, the encoder-decoder cross-entropy loss is augmented with an auxiliary CTC loss, through an extra linear projection of the encoder output representation into the output target space, to guide learning in early optimization phases when gradients aren't flowing smoothly from the decoder output to the encoder parameters due to misaligned cross-attention. The decoder cross-attention still acts over the encoder output representation maintaining the tight coupling between the encoder and decoder modules.

\section{Enforcing modularity in Seq2Seq models}
\label{sec:proposed}
Establishing an interpretable interface between the encoder and decoder components is the first step towards relaxing their tight coupling in seq2seq models. To achieve this goal, we force the encoder to output distributions over a pre-defined discrete vocabulary rather than communicating continuous vector representations to the decoder. This creates an information bottleneck \cite{bottleneck99} in the model where the decoder can communicate with the encoder's learned representations only through probability distributions over this discrete vocabulary. In addition to being interpretable, grounding the encoder outputs offers an opportunity to measure their quality independent of the decoder, if the encoder vocabulary can be mapped to the ground-truth decoder targets.

We choose an encoder output vocabulary that are sub-word units driven from the target label sequences which may deviate from the decoder output vocabulary. To force the encoder to output probabilities in the needed vocabulary space, we use the Connectionist Temporal Classification (CTC) loss. This is a novel usage of the CTC loss, not as the main loss driving the model learning process, but as a supervised function to discretize the encoder output space into a pre-defined discrete vocabulary. Even if the input-output relationship doesn't adhere to the monotonicity assumption of the CTC loss, as a module in the system, the encoder component is not expected to solve the full problem, however, the decoder module should correct any mismatch in alignment assumption through its auto-regressive generation process. 

The decoder design needs to change to cope with cross-attention over probability distributions rather than continuous hidden representations. We introduce the \texttt{AttPrep} component inside the decoder module to prepare the needed decoder internal representation for attention over the input sequence. The \texttt{AttPrep} step enables us to contain the cross-attention operation inside the decoder module. 
\begin{align*}
    Y_{1:T}^{E} &= \texttt{Encoder}(X_{1:T})\\
    \mathbf{g}_{1:T} &= \texttt{AttPrep}(Y_{1:T}^{E})\\
    y_t^{D} & \sim \texttt{Decoder}(\mathbf{g}, Y_{1:t-1}^{D})\\
    \mathcal{F}_{OBJ} &= \mathcal{F}_{CE}(Y^D, L) + \mathcal{F}_{CTC}(Y^E, L)
\end{align*}
The encoder module has a softmax normalization layer at the end so that $Y^E \in \mathbb{R}^{T\times|V_e|}$ has each row $Y^E_{i,:}$ as a probability distribution over $V_e$. For discretizing the encoder output, the CTC loss jointly optimized with the decoder cross entropy loss. Having distributions over a discrete vocabulary $V_e$ at the input of the encoder opens the space for many interesting ideas on how to harness the temporal correlations between encoder output units and common confusion patterns. We present two variants for the \texttt{AttPrep} component; the weighted embedding and beam convolution.

\subsection{Weighted Embedding \texttt{AttPrep}}
Given the encoder output distribution $Y^E$, the weighted embedding \texttt{AttPrep} (WEmb) computes an expected embedding per encoder step and combines it with sinusoidal positional encodings (PE) \cite{vaswani2017attention}, then it applies a multi-head self-attention operation (MHA) to aggregate information over all time steps.
\begin{align*}
    \mathbf{h} &= Y^E * W_{emb}\\
    \mathbf{g} &= \texttt{MHA} \left( \mathbf{h} + \texttt{PE}(\mathbf{h}) \right)
\end{align*}

\noindent
where $W_{emb}\in \mathbb{R}^{|V_e|\times d}$ and $d$ is the decoder input dimension. The first operation to compute the expected embedding is actually a 1-D time convolution operation with a receptive field of 1. It can be extended to larger receptive fields offering the opportunity to learn local confusion patterns from the encoder output. 
\begin{equation*}
    \mathbf{h} = \texttt{Conv1D}(Y^E)
\end{equation*}
One variant that we experimented with relaxes the softmax operation of the encoder output, which harshly suppresses most of the encoder output units, by applying the 1-D convolution operation above over log probabilities (WlogEmb) to allow for more information flow between encoder and decoder. 

\subsection{Beam Convolution \texttt{AttPrep}}
Rather than using the encoder output probability values, the beam convolution \texttt{AttPrep} (BeamConv) uses the rank of of the top-k hypotheses per time step. It forces a fixed bandwidth on the communication channel between the encoder and decoder, relaxing the dependence on the shape of the encoder output probability distribution. Since the top-k list doesn't preserve the unit ordering from the encoder output vector, each vocabulary unit is represented by a $p$ dimensional embedding vector. Similar to the weighted embedding \texttt{AttPrep}, a 1-D convolution operation is applied over time steps to aggregate local information followed by a multi-head self attention operation. 
\begin{align*}
    \mathbf{r} &= \texttt{Embedding} \left( \texttt{top-k}(Y^E) \right)\\
    \mathbf{h} &= \texttt{Conv1D}(\mathbf{r})\\
    \mathbf{g} &= \texttt{MHA} \left( \mathbf{h} + \texttt{PE}(\mathbf{h})\right)
\end{align*}
Where $\mathbf{r} \in \mathbb{R}^{T\times k \times p}$ with beam size $k$ and unit embedding dimension $p$.

\section{Experiments}
For our speech recognition experiments, we follow the standard Switchboard setup, with the LDC97S62 300h training set, and the Switchboard (SWB) and CallHome (CH) subsets of HUB5 Eval2000 set (LDC2002S09, LDC2000T43) for testing. Following the data preparation setup of ESPNET~\cite{espnet}, we use mean and variance normalized 83 log-mel filterbank and pitch features from 16kHz upsampled audio. As model targets, we experiment with 100 and 2000 sub-word units \cite{sentencepiece}.

We use FairSeq \cite{ott2019fairseq} for all our experiments. We use the Adam optimizer~\cite{kingma2014adam} with $eps=1e-9$ and an average batch-size of $300$ utterances. We warm-up the learning rate from $1e^{-6}$ to a peak $lr=1e^{-3}$ in $35k$ steps, keep it fixed for $1k$ steps, then linearly decrease it to $5e^{-6}$ in $44k$ steps. We follow the strong Switchboard data augmentation policy from \cite{specaugment}, but without time-warping. For inference, we don't use an external LM or joint decoding over the encoder and decoder outputs \cite{espnet}.

Our sequence-to-sequence model uses 16 transformer blocks \cite{vaswani2017attention} for the encoder and 6 for the decoder components with a convolutional context architecture \cite{ConvTransformer} where input speech features are processed using two 2-D convolution blocks with 3x3 kernels, 64 and 128 feature maps respectively, 2x2 maxpooling, and ReLU non-linearity. Both encoder and decoder transformer blocks have 1024 dimensions, 16 heads, and 4096 dimensional feed-forward network. Sinusoidal positional embeddings are added to the output of the encoder 2-D convolutional context layers. 
\begin{table}[H]
\centering
\caption{Baseline Seq2Seq ASR Models}
\label{baseline-table}
\resizebox{\linewidth}{!}{
\begin{tabular}{c|c|c|c|c|c|c}
\toprule
\multicolumn{2}{c|}{BPE Units} & Beam & \multicolumn{2}{c|}{Loss Criterion} & \multicolumn{2}{c}{Eval 2000}  \\
CTC & CE & Size & CTC & CE & SWB   & CH     \\
\toprule
\multicolumn{7}{c}{Our baseline implementation} \\
\midrule
100 & - & 1 & \cmark & \xmark & 11.2 & 21.7 \\
- & 100 & 20 & \xmark & \cmark & 9.6 & 19.6 \\
100 & 100 & 20 & \cmark & \cmark & 8.5 & 17.0 \\
\midrule
2000 & - & 1 & \cmark & \xmark & 12.6 & 24.3 \\
- & 2000 & 20 & \xmark & \cmark & 8.5 & 17.7 \\
2000 & 2000 & 20 & \cmark & \cmark & 8.5 & 18.0 \\
\midrule
100 & 2000 & 20 & \cmark & \cmark & 7.8 & 17.7 \\
\toprule
\multicolumn{7}{c}{LAS + SpecAugment \cite{specaugment}} \\
\midrule
- & 1000 & 8 & \xmark & \cmark & 7.3 & 14.4 \\
\toprule
\multicolumn{7}{c}{ESPNET \cite{Karita2019ACS}}  \\
\midrule
2000 & 2000 & 20 & \cmark & \cmark & 9.0 & 18.1 \\
\toprule
\multicolumn{7}{c}{Kaldi Hybrid system \footnotemark \cite{povey2016purely}}\\
\midrule
- & - & - & \multicolumn{2}{c|}{LF-MMI} & 8.8 & 18.1 \\
\bottomrule
\end{tabular}
}
\end{table}
\footnotetext{Results from Kaldi's recent best recipe on GitHub}

\subsection{Baseline models performance}
Table \ref{baseline-table} shows the word error rates (WER) of our ASR baseline implementations employing the three approaches for seq2seq modeling, along with the current SOTA systems in the literature. In line with \citet{choicebpe}, the auto-regressive encoder-decoder models benefits from larger modeling units as opposed to the CTC-optimized one that works best with shorter linguistic units. Encoder-decoder models trained by joint optimization of the CTC and cross-entropy losses benefits from a hybrid setup with two different vocabulary sets.

The problem of tight coupling of the encoder and decoder components in the seq2seq model is highlighted in table \ref{baseline-entanglement}. The decoder cross-attention over the encoder hidden representation makes it not only conditioned on the encoder outputs but also dependent on the encoder architectural decisions and internal hidden representations. The whole ASR system fall apart under the interchangability test, i.e. switching an encoder with another similar one that is only different in its initial random seed, which brings our point about the lack of modularity in encoder-decoder models.

The same level of coupling is also observed in models that utilizes the auxiliary CTC loss to accelerate early learning stages \cite{suyounkim, Karita2019ACS}.
\begin{table}[H]
\centering
\caption{Effect of switching encoders on WER}
\label{baseline-entanglement}
\resizebox{\linewidth}{!}{
\begin{tabular}{l|c|c|c|c}
\toprule
Random & \multicolumn{2}{c|}{Loss Criterion} & \multicolumn{2}{c}{Eval2000} \\
Seed & CTC & CE & SWB   & CH     \\
\midrule
Seed 1 & \xmark & \cmark & 8.5  & 17.7         \\
Seed 2 & \xmark & \cmark & 8.5  & 18.3         \\
\midrule
Enc. Swap 1 & \xmark & \cmark & 569.0  & 892.0\\
Enc. Swap 2 & \xmark & \cmark & 597.2  & 759.3\\
\midrule
\midrule
Seed 1 & \cmark & \cmark & 7.8 & 17.7\\
Seed 2 & \cmark & \cmark & 7.9 & 17.1\\
\midrule
Enc. Swap 1 & \cmark & \cmark & 850.5 & 942.8 \\
Enc. Swap 2 & \cmark & \cmark & 747.4 & 1094.5\\
\bottomrule
\end{tabular}
}
\end{table}
\subsection{Performance of the proposed modular Seq2Seq models}
\label{sec:decouple_performance}
\begin{table*}[ht!]
\centering
\caption{Evidence of modularity using our proposed decoupling techniques in attention based ASR Systems}
\label{tab:model_swap}
\resizebox{\textwidth}{!}{
\begin{tabular}{l|c|c|c|c|l|c|c|c|c||c|c|c|c}
\toprule
\multicolumn{5}{c|}{Model 1} & \multicolumn{5}{c||}{Model 2} & \multicolumn{2}{c|}{Enc2 Dec1} & \multicolumn{2}{c}{Enc1 Dec2} \\
Architecture & RF & K & SWB & CH & Architecture & RF & K & SWB & CH & SWB   & CH & SWB & CH \\
\midrule
BeamConv & 1 & 20 & 8.4 & 17.6 & BeamConv & 1 & 20 & 8.6 & 17.6  & 8.6 & 17.8 & 8.5 & 17.3 \\
BeamConv & 3 & 10 & 8.7 & 17.3  & BeamConv & 3 & 10 & 9.2 & 17.2  & 9.7 & 17.3 & 8.9 & 17.5 \\
WEmb     & 3 &  - & 8.7 & 17.9 & WEmb     & 3 &  - & 8.8 & 18.3  &  9.0  & 18.3 & 8.6 & 18.0 \\
WLogEmb  & 3 &  - & 8.0 & 16.4 & WLogEmb  & 3 &  - & 8.0 & 16.4  & 61.4 & 60.1 & 55.2 & 62.7 \\
\midrule
\midrule
BeamConv & 3 & 20 & 8.8 & 17.3 & BeamConv & 1 & 10 & 8.3 & 17.6  & 8.4 & 17.4 & 9.3 & 17.4 \\
WEmb &3 & - & 8.7  & 17.9 & BeamConv & 1 & 10 & 8.3 & 17.6 & 8.4 & 17.2 & 8.7 & 18.4 \\
WEmb & 3 & - & 8.7 & 17.9 & WLogEmb & 1 & - & 7.8 & 17.4 & 8.7 & 17.9     & 96.3 & 265.1 \\
BeamConv & 1 & 20 & 8.4 & 17.5 & WLogEmb &3 & - & 8.0 & 16.4 & 8.3 & 17.3 & 85.0 & 112.5 \\
\bottomrule
\end{tabular}
}
\end{table*}

Tables \ref{tab:model_swap} and \ref{tab:beamconv} show that the proposed modular seq2seq models are competitive with SOTA performance levels \ref{baseline-table}, and that the models are highly modular. Performance does not degrade when exchanging encoders and decoders trained from different initial seeds or choices of architectures.

The information bottleneck at the encoder outputs
is critical for this result, as shown from the WLogEmb architecture. Relaxing the bandwidth constraint on the encoder-decoder connection by utilizing the log distribution lets the decoder rely on specific patterns of errors at the tail of the encoder output distribution for its final decisions. This improves the overall performance but breaks modularity when a different encoder is used, as shown in table \ref{tab:model_swap}.
\begin{table}[H]
\centering
\caption{WER of the encoder and decoder outputs for the proposed modular models}
\label{tab:beamconv}
\resizebox{\linewidth}{!}{
\begin{tabular}{l|c|c|c|c|c|c}
\toprule
\multirow{2}{*}{Architecture} & \multirow{2}{*}{RF} & Top&\multicolumn{2}{c|}{Encoder} & \multicolumn{2}{c}{Decoder}\\
    & & k & SWB & CH & SWB   & CH \\
\midrule
WEmb & 1  & - & 12.2 & 23.1 & 8.7 & 18.0 \\
WEmb & 3  & - & 11.9 & 22.6 & 8.7 & 17.9 \\
WEmb & 5  & - & 12.0 & 23.0 & 8.4 & 17.9 \\
\midrule
WLogEmb & 1 & - & 11.1 & 21.9 & 7.8  & 17.2 \\
WLogEmb & 3 & - & 11.1 & 21.7 & 8.0  & 16.4 \\
WLogEmb & 5 & - & 10.9 & 22.2 & 7.7  & 16.3 \\
\midrule
BeamConv & 1 & 10  & 11.0 & 21.7 & 8.3  & 17.6 \\
BeamConv & 1 & 20  & 11.2 & 21.9 & 8.4  & 17.6 \\
BeamConv & 1 & 50  & 11.2 & 21.7 & 8.4  & 17.2 \\
BeamConv & 3 & 10  & 11.2 & 21.9 & 8.7  & 17.3 \\
BeamConv & 3 & 20  & 10.8 & 22.0 & 8.8  & 17.3 \\
BeamConv & 3 & 50  & 11.5 & 21.7 & 9.9  & 17.0 \\
\bottomrule
\end{tabular}
}
\end{table}
\subsection{Advantages of Modularity}
\label{sec:use}
Building modular seq2seq models brings about many advantages including interpretable modular interface, functional independence of each module, and interchangeability. In addition to opening up the space for designing modules implementing specific functions, grounding each module's output into some interpretable discrete vocabulary allows for debugging and measure the quality of each modular component in the system. In our speech recognition experiments, the encoder acts like an acoustic model by mapping input acoustic evidences into low level linguistic units, while the decoder, acting like a language model, aggregates distributions of such units to generate the most likely full sentence. For example, table \ref{tab:beamconv} shows how the overall performance is improving going from the encoder to the decoder module in the last two columns. 

Another benefit of modular independence, which we enjoy in software design but not in building fully-differentiable seq2seq models, is the ability to carefully build one critical module to higher levels of performance then switch it into the full system without the need for any model fine-tuning. The new module may reflect a new architecture design in that module, e.g. from LSTMs to Transformers, or simply more training data that become available for that module. Such ``modular upgrade'' capability is demonstrated in table \ref{tab:modular_upgrade} where the upgraded encoder performance is reflected into the overall system WER. In our case, we just used an encoder model that is trained independently using the CTC loss for a larger number of updates.

\begin{table}[H]
\centering
\caption{WER without (scratched) and with modular upgrade of the decoupled model.}
\label{tab:modular_upgrade}
\resizebox{\linewidth}{!}{
\begin{tabular}{l|c|c|c|c|c|c}
\toprule
\multirow{2}{*}{Architecture} & \multirow{2}{*}{RF} & Top&\multicolumn{2}{c|}{Original Enc} & \multicolumn{2}{c}{Decoder}\\
    & & k & SWB & CH & SWB   & CH \\
\midrule
\multicolumn{3}{l|}{Upgraded Encoder} & 11.2 & 21.1 & - & - \\
\midrule
BeamConv & 1 & 10  & 11.0 & 21.7 & \sout{8.3} 8.7  & \sout{17.6} 16.6 \\
BeamConv & 3 & 50  & 11.5 & 21.7 & \sout{9.9} 9.3  & \sout{17.0} 16.4 \\
\midrule
WEmb & 1 & - & 12.2 & 23.1 & \sout{8.7} 8.9 & \sout{18.0} 17.8 \\
WEmb & 5 & - & 12.0 & 23.0 & \sout{8.4} 8.8 & \sout{17.9} 17.4 \\
\bottomrule
\end{tabular}
}
\end{table}

Table \ref{tab:postedit} presents experiments for a slightly different scenario where one module is trained from scratch conditioned on the output of its parent module with frozen parameters, dubbed \texttt{PostEdit} in our experiments. The beam convolution attention preparation architecture, which uses only the rank of encoder hypotheses rather than probability values, shows much more resilience and ability to fix frozen parent module errors compared to the weighed embedding architecture. There is still a slight degradation of the final decoder performance when trained conditionally on the encoder output -- without joint fine-tuning. The reason for that is the lack of data augmentation effect when training the decoder module, as a side effect of modular components, because the encoder is trained to be invariant to augmentation when producing its final probability distribution. This can be treated by designing data augmentation techniques suitable to be applied at the input of each module, which we refer to future work.
\begin{table}[H]
\centering
\caption{\texttt{PostEdit} conditional training of the Decoder module}
\label{tab:postedit}
\resizebox{\linewidth}{!}{
\begin{tabular}{l|c|c|c|c|c|c}
\toprule
\multirow{2}{*}{Architecture} & \multirow{2}{*}{RF} & Top&\multicolumn{2}{c|}{Encoder} & \multicolumn{2}{c}{PostEdit Dec.}\\
    & & k & SWB & CH & SWB   & CH \\
\midrule
BeamConv & 1 & 10  & 11.2 & 21.7 & 9.5  & 18.6 \\
BeamConv & 3 & 50  & 11.2 & 21.7 & 10.6  & 17.8 \\
\midrule
WEmb & 1 & - & 11.2 & 21.7 & 17.3  & 27.4 \\
WEmb & 5 & - & 11.2 & 21.7 & 12.2  & 17.9 \\
\bottomrule
\end{tabular}
}
\end{table}
Modularity provides us with the ability to create an ensemble of exponential number of models, e.g by training 3 different modular seq2seq systems, we end up with an ensemble of 9. In table \ref{tab:ensemble} we show that a modular ensemble of 4 provides further improvement over the WER of an ensemble of the original 2 models.
\begin{table}[H]
\centering
\caption{Modular Ensembling further improves WER}
\label{tab:ensemble}
\resizebox{\linewidth}{!}{
\begin{tabular}{l|c|c|c|c}
\toprule
\multirow{2}{*}{Architecture} & \multirow{2}{*}{RF} & Top &\multicolumn{2}{c}{Decoder}\\
     & & k & SWB   & CH \\
\midrule
BeamConv & 1 & 10 & 8.3  & 17.6 \\
WEmb & 3 & - & 8.7  & 18.0 \\
\midrule
Ensemble of 2 & 1/3 & 10/- & 7.7  & 16.1 \\
+ 2 using modular swap & 1/3 & 10/-  & 7.7  & 15.8 \\
\bottomrule
\end{tabular}
}
\end{table}

\section{Related work}
\label{sec:prior}
This work is applying the component modularity notion from the design and analysis of complex systems \cite{Baldwin_99} to fully-differentiable seq2seq models which achieved impressive levels of performance across many tasks \cite{chan2016listen,bahdanau2016end, bahdanau2014neural, vaswani2017attention}. The Connectionist Temporal Classification (CTC) loss \cite{graves2006connectionist} was applied as a sequence level loss for training encoder-only speech recognition models \cite{Graves_e2e, Hannun2014}, and as a joint loss in attention-based systems for encouraging monotonic alignment between input and output sequences \cite{suyounkim}. The CTC loss serves the purpose of introducing an information bottleneck \cite{bottleneck99} through discretizing the encoder output into an interpretable vocabulary space.

By enforcing modularity between the encoder and decoder components in seq2seq models, the decoder module can be viewed as a post-edit module to the recognition output of the encoder. Also, the decoder can be viewed as an instance of a differentiable beam-search decoder \cite{diff_dec_19}.

There is a long history of research in learning disentangled, distributed hidden representations \cite{Rumelhart, hinton1986learning}, unsupervised discovery of abstract factors of variations within the training data \cite{Bengio_rep_learning, yann_rep_learning, chen2016infogan, burgess2018understanding}. This line of research is complementary to our work which enforces modularity only at the link connecting two big components in a seq2seq system. In this work, a component is defined as a deep and complex network with multiple layers of representations which serves a specific function within the bigger system, and outputs distributions over interpretable vocabulary units. 

Another line of research that is related to ours centers around inducing a modular structure on the space of learned concepts through hierarchically gating information flow or via high-level concept blueprints \cite{andreas2016neural, devin16, ranzato19} to enable zero and few-shot transfer learning \cite{andreas2017modular, socher2013zero}, multi-lingual and cross-lingual learning \cite{adams2019massively, dalmia2018sequence, swietojanski2012unsupervised}. 

Hybrid HMM-DNN speech recognition systems \cite{Gales_2007, asr_2012} are modular by design but they lack end-to-end learning capability. We aim at bringing the same modular properties without losing quality nor full differentiability.

\section{Conclusion}
\label{sec:conc}
Motivated by modular software and system design literature, we presented a method for inducing modularity in attention-based seq2seq models through discretizing the encoder output into a real-world vocabulary units. The Connectionist Temporal Classification (CTC) loss is applied to the encoder outputs to ground them into the predefined vocabulary while respecting their sequential nature. The learned model adhere to the three properties of modular systems -- independence, interchangeability, and clearness of interface -- while achieving a competitive WER performance in the standard 300h Switchboard task of $8.3\%$ and $17.6\%$ on the SWB and CH subsets respectively. Our future work focuses on extending this work to other sequence-to-sequence machine translation and language processing tasks, as well as exploring the benefits of modular transfer in multi-task and multi-modal settings. 

\section{Acknowledgements}
\label{sec:ack}
The authors would like to thank Paul Michel, Dmytro Okhonko, Matthew Weisner for their helpful discussions and comments.

\bibliography{acl2020}

\begin{thebibliography}{37}
\expandafter\ifx\csname natexlab\endcsname\relax\def\natexlab#1{#1}\fi

\bibitem[{Adams et~al.(2019)Adams, Wiesner, Watanabe, and
  Yarowsky}]{adams2019massively}
Oliver Adams, Matthew Wiesner, Shinji Watanabe, and David Yarowsky. 2019.
\newblock {Massively Multilingual Adversarial Speech Recognition}.
\newblock In \emph{Proc. NAACL-HLT}.

\bibitem[{Andreas et~al.(2017)Andreas, Klein, and Levine}]{andreas2017modular}
Jacob Andreas, Dan Klein, and Sergey Levine. 2017.
\newblock {Modular multitask Reinforcement Learning with Policy Sketches}.
\newblock In \emph{Proc. ICML}.

\bibitem[{Andreas et~al.(2016)Andreas, Rohrbach, Darrell, and
  Klein}]{andreas2016neural}
Jacob Andreas, Marcus Rohrbach, Trevor Darrell, and Dan Klein. 2016.
\newblock {Neural Module Networks}.
\newblock In \emph{Proc. CVPR}.

\bibitem[{Bahdanau et~al.(2015)Bahdanau, Cho, and Bengio}]{bahdanau2014neural}
Dzmitry Bahdanau, Kyunghyun Cho, and Yoshua Bengio. 2015.
\newblock {Neural Machine Translation by Jointly Learning to Align and
  Translate}.
\newblock \emph{Proc. ICLR}.

\bibitem[{Bahdanau et~al.(2016)Bahdanau, Chorowski, Serdyuk, Brakel, and
  Bengio}]{bahdanau2016end}
Dzmitry Bahdanau, Jan Chorowski, Dmitriy Serdyuk, Philemon Brakel, and Yoshua
  Bengio. 2016.
\newblock {End-to-End Attention-based Large Vocabulary Speech Recognition}.
\newblock In \emph{Proc. ICASSP}.

\bibitem[{Baldwin and Clark(1999)}]{Baldwin_99}
Carliss~Y. Baldwin and Kim~B. Clark. 1999.
\newblock \emph{{Design Rules: The Power of Modularity Volume 1}}.
\newblock MIT Press.

\bibitem[{Bengio(2013)}]{Bengio_rep_learning}
Yoshua Bengio. 2013.
\newblock Deep learning of representations: Looking forward.
\newblock \emph{CoRR}.

\bibitem[{Chan et~al.(2016)Chan, Jaitly, Le, and Vinyals}]{chan2016listen}
William Chan, Navdeep Jaitly, Quoc Le, and Oriol Vinyals. 2016.
\newblock {Listen, Attend and Spell: A neural network for large vocabulary
  conversational speech recognition}.
\newblock In \emph{Proc. ICASSP}.

\bibitem[{Chen et~al.(2016)Chen, Duan, Houthooft, Schulman, Sutskever, and
  Abbeel}]{chen2016infogan}
Xi~Chen, Yan Duan, Rein Houthooft, John Schulman, Ilya Sutskever, and Pieter
  Abbeel. 2016.
\newblock {InfoGAN: Interpretable Representation Learning by Information
  Maximizing Generative Adversarial Nets}.
\newblock In \emph{Proc. NeurIPS}.

\bibitem[{Collobert et~al.()Collobert, Hannun, and Synnaeve}]{diff_dec_19}
Ronan Collobert, Awni Hannun, and Gabriel Synnaeve.
\newblock A fully differentiable beam search decoder.
\newblock In \emph{ICML 2019}.

\bibitem[{Dalmia et~al.(2018)Dalmia, Sanabria, Metze, and
  Black}]{dalmia2018sequence}
Siddharth Dalmia, Ramon Sanabria, Florian Metze, and Alan~W Black. 2018.
\newblock {Sequence-Based Multi-Lingual Low Resource Speech Recognition}.
\newblock In \emph{Proc. ICASSP}.

\bibitem[{Devin et~al.(2016)Devin, Gupta, Darrell, Abbeel, and
  Levine}]{devin16}
Coline Devin, Abhishek Gupta, Trevor Darrell, Pieter Abbeel, and Sergey Levine.
  2016.
\newblock Learning modular neural network policies for multi-task and
  multi-robot transfer.
\newblock \emph{CoRR}.

\bibitem[{Gales and Young()}]{Gales_2007}
Mark Gales and Steve Young.
\newblock The application of hidden markov models in speech recognition.
\newblock \emph{Found. Trends Signal Process.}, 1(3).

\bibitem[{Graves et~al.(2006)Graves, Fern{\'a}ndez, Gomez, and
  Schmidhuber}]{graves2006connectionist}
Alex Graves, Santiago Fern{\'a}ndez, Faustino Gomez, and J{\"u}rgen
  Schmidhuber. 2006.
\newblock {Connectionist Temporal Classification: Labelling Unsegmented
  Sequence Data with Recurrent Neural Networks}.
\newblock In \emph{Proc. ICML}.

\bibitem[{Graves and Jaitly(2014)}]{Graves_e2e}
Alex Graves and Navdeep Jaitly. 2014.
\newblock Towards end-to-end speech recognition with recurrent neural networks.
\newblock In \emph{Proceedings of the 31st International Conference on
  International Conference on Machine Learning}.

\bibitem[{Hannun et~al.(2014)Hannun, Case, Casper, Catanzaro, Diamos, Elsen,
  Prenger, Satheesh, Sengupta, Coates, and Ng}]{Hannun2014}
Awni~Y. Hannun, Carl Case, Jared Casper, Bryan Catanzaro, Greg Diamos, Erich
  Elsen, Ryan Prenger, Sanjeev Satheesh, Shubho Sengupta, Adam Coates, and
  Andrew~Y. Ng. 2014.
\newblock Deep speech: Scaling up end-to-end speech recognition.
\newblock \emph{ArXiv}.

\bibitem[{Higgins et~al.(2017)Higgins, Matthey, Pal, Burgess, Glorot,
  Botvinick, Mohamed, and Lerchner}]{burgess2018understanding}
Irina Higgins, Loic Matthey, Arka Pal, Christopher Burgess, Xavier Glorot,
  Matthew Botvinick, Shakir Mohamed, and Alexander Lerchner. 2017.
\newblock {beta-VAE: Learning Basic Visual Concepts with a Constrained
  Variational Framework}.
\newblock In \emph{Proc. ICLR}.

\bibitem[{{Hinton} et~al.(2012){Hinton}, {Deng}, {Yu}, {Dahl}, {Mohamed},
  {Jaitly}, {Senior}, {Vanhoucke}, {Nguyen}, {Sainath}, and
  {Kingsbury}}]{asr_2012}
G.~{Hinton}, L.~{Deng}, D.~{Yu}, G.~E. {Dahl}, A.~{Mohamed}, N.~{Jaitly},
  A.~{Senior}, V.~{Vanhoucke}, P.~{Nguyen}, T.~N. {Sainath}, and
  B.~{Kingsbury}. 2012.
\newblock Deep neural networks for acoustic modeling in speech recognition: The
  shared views of four research groups.
\newblock \emph{IEEE Signal Processing Magazine}, 29(6).

\bibitem[{Hinton et~al.(1986)}]{hinton1986learning}
Geoffrey~E. Hinton et~al. 1986.
\newblock Learning distributed representations of concepts.
\newblock In \emph{Proceedings of the eighth annual conference of the cognitive
  science society}.

\bibitem[{Irie et~al.(2019)Irie, Prabhavalkar, Kannan, Bruguier, Rybach, and
  Nguyen}]{choicebpe}
Kazuki Irie, Rohit Prabhavalkar, Anjuli Kannan, Antoine Bruguier, David Rybach,
  and Patrick Nguyen. 2019.
\newblock {On the Choice of Modeling Unit for Sequence-to-Sequence Speech
  Recognition}.
\newblock In \emph{Proc. InterSpeech}.

\bibitem[{Karita et~al.(2019)Karita, Chen, Hayashi, and other}]{Karita2019ACS}
Shigeki Karita, Nanxin Chen, Tomoki Hayashi, and other. 2019.
\newblock {A Comparative Study on Transformer vs RNN in Speech Applications}.
\newblock \emph{Proc. ASRU}.

\bibitem[{Kim et~al.(2017)Kim, Hori, and Watanabe}]{suyounkim}
Suyoun Kim, Takaaki Hori, and Shinji Watanabe. 2017.
\newblock {Joint CTC-Attention based End-to-End Speech Recognition using
  Multi-task Learning}.
\newblock In \emph{Proc. ICASSP}.

\bibitem[{Kingma and Ba(2014)}]{kingma2014adam}
Diederik~P Kingma and Jimmy~Lei Ba. 2014.
\newblock {Adam: A Method for Stochastic Optimization}.
\newblock In \emph{Proc. ICLR}.

\bibitem[{Kudo and Richardson(2018)}]{sentencepiece}
Taku Kudo and John Richardson. 2018.
\newblock {SentencePiece: A simple and language independent subword tokenizer
  and detokenizer for neural text processing}.
\newblock In \emph{Proc. EMNLP: System Demonstrations}.

\bibitem[{Mathieu et~al.(2016)Mathieu, Zhao, Zhao, Ramesh, Sprechmann, and
  LeCun}]{yann_rep_learning}
Michael~F Mathieu, Junbo~Jake Zhao, Junbo Zhao, Aditya Ramesh, Pablo
  Sprechmann, and Yann LeCun. 2016.
\newblock Disentangling factors of variation in deep representation using
  adversarial training.
\newblock In \emph{Advances in Neural Information Processing Systems 29}.

\bibitem[{Mohamed et~al.(2019)Mohamed, Okhonko, and
  Zettlemoyer}]{ConvTransformer}
Abdelrahman Mohamed, Dmytro Okhonko, and Luke Zettlemoyer. 2019.
\newblock {Transformers with convolutional context for ASR}.
\newblock In \emph{arXiv preprint arXiv:1904.11660}.

\bibitem[{Ott et~al.(2019)Ott, Edunov, Baevski, Fan, Gross, Ng, Grangier, and
  Auli}]{ott2019fairseq}
Myle Ott, Sergey Edunov, Alexei Baevski, Angela Fan, Sam Gross, Nathan Ng,
  David Grangier, and Michael Auli. 2019.
\newblock {fairseq: A Fast, Extensible Toolkit for Sequence Modeling}.
\newblock In \emph{Proc. NAACL-HLT: Demonstrations}.

\bibitem[{Park et~al.(2019)Park, Chan, Zhang, Chiu, Zoph, Cubuk, and
  Le}]{specaugment}
Daniel~S Park, William Chan, Yu~Zhang, Chung-Cheng Chiu, Barret Zoph, Ekin~D
  Cubuk, and Quoc~V Le. 2019.
\newblock {SpecAugment: A simple data augmentation method for automatic speech
  recognition}.
\newblock In \emph{Proc. Interspeech}.

\bibitem[{Povey et~al.(2016)Povey, Peddinti, Galvez, Ghahremani, Manohar, Na,
  Wang, and Khudanpur}]{povey2016purely}
Daniel Povey, Vijayaditya Peddinti, Daniel Galvez, Pegah Ghahremani, Vimal
  Manohar, Xingyu Na, Yiming Wang, and Sanjeev Khudanpur. 2016.
\newblock Purely sequence-trained neural networks for asr based on lattice-free
  mmi.

\bibitem[{Purushwalkam et~al.(2019)Purushwalkam, Nickel, Gupta, and
  Ranzato}]{ranzato19}
Senthil Purushwalkam, Maximilian Nickel, Abhinav Gupta, and Marc'Aurelio
  Ranzato. 2019.
\newblock Task-driven modular networks for zero-shot compositional learning.
\newblock \emph{CoRR}.

\bibitem[{Rumelhart et~al.(1986)Rumelhart, McClelland, and PDP
  Research~Group}]{Rumelhart}
David~E. Rumelhart, James~L. McClelland, and CORPORATE PDP Research~Group,
  editors. 1986.
\newblock \emph{Parallel Distributed Processing: Explorations in the
  Microstructure of Cognition, Vol. 1: Foundations}.

\bibitem[{Socher et~al.(2013)Socher, Ganjoo, Manning, and Ng}]{socher2013zero}
Richard Socher, Milind Ganjoo, Christopher~D Manning, and Andrew Ng. 2013.
\newblock {Zero-Shot Learning Through Cross-Modal Transfer}.
\newblock In \emph{Proc. NeurIPS}.

\bibitem[{Sutskever et~al.(2014)Sutskever, Vinyals, and
  Le}]{sutskever2014sequence}
Ilya Sutskever, Oriol Vinyals, and Quoc~V Le. 2014.
\newblock Sequence to sequence learning with neural networks.
\newblock In \emph{Advances in neural information processing systems}, pages
  3104--3112.

\bibitem[{Swietojanski et~al.(2012)Swietojanski, Ghoshal, and
  Renals}]{swietojanski2012unsupervised}
Pawel Swietojanski, Arnab Ghoshal, and Steve Renals. 2012.
\newblock {Unsupervised cross-lingual knowledge transfer in DNN-based LVCSR}.
\newblock In \emph{Proc. SLT}.

\bibitem[{Tishby et~al.(1999)Tishby, Pereira, and Bialek}]{bottleneck99}
Naftali Tishby, Fernando~C. Pereira, and William Bialek. 1999.
\newblock The information bottleneck method.
\newblock In \emph{Proc. of the 37-th Annual Allerton Conference on
  Communication, Control and Computing}.

\bibitem[{Vaswani et~al.(2017)Vaswani, Shazeer, Parmar, Uszkoreit, Jones,
  Gomez, Kaiser, and Polosukhin}]{vaswani2017attention}
Ashish Vaswani, Noam Shazeer, Niki Parmar, Jakob Uszkoreit, Llion Jones,
  Aidan~N Gomez, {\L}ukasz Kaiser, and Illia Polosukhin. 2017.
\newblock {Attention is all you need}.
\newblock In \emph{Proc. NeurIPS}.

\bibitem[{Watanabe et~al.(2018)Watanabe, Hori, Karita, Hayashi, Nishitoba, Unno
  et~al.}]{espnet}
Shinji Watanabe, Takaaki Hori, Shigeki Karita, Tomoki Hayashi, Jiro Nishitoba,
  Yuya Unno, et~al. 2018.
\newblock {ESPnet: End-to-End Speech Processing Toolkit}.
\newblock In \emph{Proc. InterSpeech}.

\end{thebibliography}
\bibliographystyle{acl_natbib}

\end{document}